\documentclass[10pt, conference, compsocconf]{IEEEtran}
\ifCLASSINFOpdf
\else
\fi

\usepackage{times}
\usepackage{epsfig}
\usepackage{graphicx}
\usepackage{amsmath}
\usepackage{amssymb}

\usepackage{amsfonts}
\usepackage{etoolbox}
\usepackage{multirow}
\usepackage{makecell}
\usepackage{hyperref}
\usepackage{tabularx}
\usepackage{multicol}

\usepackage[utf8]{inputenc} % usually not needed (loaded by default)
\usepackage[T1]{fontenc}

\usepackage{xcolor}
\usepackage[normalem]{ulem}
\definecolor{tangelo}{rgb}{0.98, 0.3, 0.0}

\newcommand{\nbNegTree}{750}
\newcommand{\HomoImgToRef}{$H^r$}
\newcommand{\HomoRefToImg}{$H^i$}

\newcommand{\RefFrame}{$R$}

\newcommand{\NbNearestNeighbor}{$\alpha$}
\newcommand{\NearestNeighborMatchingRatio}{$\rho$}
\newcommand{\Score}{$g$}
\newcommand{\DescVect}{$v$}

\newcommand{\BlurSigma}{$\sigma$}
\newcommand{\Downsize}{$\phi$}
\newcommand{\NbKp}{$\gamma$}
\makeatletter
\patchcmd{\@makecaption}
  {\scshape}
  {}
  {}
  {}
\makeatother

\usepackage[printonlyused]{acronym}
\acrodef{SIFT}{Scale Invariant Feature Transform}
\acrodef{SURF}{Speeded Up Robust Features}

\acrodef{CBIR}{Content-Based Image Retrieval}
\acrodef{CNN}{Convolutional Neural Network}
\acrodef{VPR}{Visual Place Recognition}
\acrodef{SFM}{Structure-From Motion}
\acrodef{MVS}{Multi-View Stereo}
\acrodef{HSV}{Hue-Saturation-Value}
\acrodef{PCA}{Principal Component Analysis}
\acrodef{Person Re-Id}{Person Re-Identification}
\acrodef{Re-Id}{Re-Identification}

\acrodef{TF}{Term Frequency}
\acrodef{IDF}{Inverse Document Frequency}

\acrodef{ROC}{Receiver Operating Characteristic}
\acrodef{AUC}{Area Under the Curve}
\acrodef{mAP}{mean Average Precision}
\acrodef{AP}{Average Precision}
\acrodef{F1}{Harmonic Average of the Precision and Recall}
\acrodef{PR}{Precision Recall}
\acrodef{P@K}{Precision at rank K}
\acrodef{R@K}{Recall at rank K}
\acrodef{P@1}{Precision at rank 1}
\acrodef{R-P}{R Precision}

\acrodef{BoW}{Bag of Words}
\acrodef{GV}{Geometric Verification}
\acrodef{LR}{Lowe Ratio}
\acrodef{LBP}{Local Binary Patterns}

\acrodef{DBH}{Diameter at Breast Height}
\acrodef{RFID}{Radio Frequency Identification}

\acrodef{RP}{Red Pine}
\acrodef{EL}{Elm}

\renewcommand{\thetable}{\arabic{table}}

\title{\LARGE \bf
Tree bark re-identification using a deep-learning feature descriptor
}

\author{\IEEEauthorblockN{Martin Robert, Patrick Dallaire, Philippe Gigu\`ere}
\IEEEauthorblockA{Department of Computer Science and Software Engineering\\
Université Laval, Quebec City, Canada\\
martin.robert.3@ulaval.ca, \{philippe.giguere,patrick.dallaire\}@ift.ulaval.ca}
}

\begin{document}

% conference papers do not typically use \thanks and this command
% is locked out in conference mode. If really needed, such as for
% the acknowledgment of grants, issue a \IEEEoverridecommandlockouts
% after \documentclass

% for over three affiliations, or if they all won't fit within the width
% of the page, use this alternative format:
% 
%\author{\IEEEauthorblockN{Michael Shell\IEEEauthorrefmark{1},
%Homer Simpson\IEEEauthorrefmark{2},
%James Kirk\IEEEauthorrefmark{3}, 
%Montgomery Scott\IEEEauthorrefmark{3} and
%Eldon Tyrell\IEEEauthorrefmark{4}}
%\IEEEauthorblockA{\IEEEauthorrefmark{1}School of Electrical and Computer Engineering\\
%Georgia Institute of Technology,
%Atlanta, Georgia 30332--0250\\ Email: see http://www.michaelshell.org/contact.html}
%\IEEEauthorblockA{\IEEEauthorrefmark{2}Twentieth Century Fox, Springfield, USA\\
%Email: homer@thesimpsons.com}
%\IEEEauthorblockA{\IEEEauthorrefmark{3}Starfleet Academy, San Francisco, California 96678-2391\\
%Telephone: (800) 555--1212, Fax: (888) 555--1212}
%\IEEEauthorblockA{\IEEEauthorrefmark{4}Tyrell Inc., 123 Replicant Street, Los Angeles, California 90210--4321}}

% use for special paper notices
%\IEEEspecialpapernotice{(Invited Paper)}

% make the title area
\maketitle

%%%%%%%%%%%%%%%%%%%%%%%%%%%%%%%%%%% ABSTRACT %%%%%%%%%%%%%%%%%%%%%%%%%%%%%%%%%%%%%%%%%%%%%

\begin{abstract}
The ability to visually re-identify objects is a fundamental capability in vision systems. Oftentimes, it relies on collections of visual signatures based on descriptors, such as \acs{SIFT} or \acs{SURF}. However, these traditional descriptors were designed for a certain domain of surface appearances and geometries (limited relief). Consequently, highly-textured surfaces such as tree bark pose a challenge to them. In turn, this makes it more difficult to use trees as identifiable landmarks for navigational purposes (robotics) or to track felled lumber along a supply chain (logistics). We thus propose to use data-driven descriptors trained on bark images for tree surface re-identification. To this effect, we collected a large dataset containing 2,400 bark images with strong illumination changes, annotated by surface and with the ability to pixel-align them. We used this dataset to sample from more than 2 million $64 \times 64$ pixel patches to train our novel local descriptors \texttt{DeepBark} and \texttt{SqueezeBark}. Our \texttt{DeepBark} method has shown a clear advantage against the hand-crafted descriptors SIFT and SURF. For instance, we demonstrated that \texttt{DeepBark} can reach a mAP of 87.2\% when retrieving 11 relevant bark images, i.e. corresponding to the same physical surface, to a bark query against 7,900 images. Our work thus suggests that re-identifying tree surfaces in a challenging illuminations context is possible. We also make public our dataset, which can be used to benchmark surface re-identification techniques.
\end{abstract}


% For peer review papers, you can put extra information on the cover
% page as needed:
% \ifCLASSOPTIONpeerreview
% \begin{center} \bfseries EDICS Category: 3-BBND \end{center}
% \fi
%
% For peerreview papers, this IEEEtran command inserts a page break and
% creates the second title. It will be ignored for other modes.
\IEEEpeerreviewmaketitle

%%%%%%%%%%%%%%%%%%%%%%%%%%%%%%%%%%%% INTRODUCTION %%%%%%%%%%%%%%%%%%%%%%%%%%%%%%%%%%%%
\section{INTRODUCTION} % https://cs.stanford.edu/people/widom/paper-writing.html
The tracking of objects is important in many domains. For instance, tracking within the supply chain is a key element of the Industry 4.0 philosophy~\cite{Hinkka2012}. In the forestry industry, it would consist in re-identifying trees cut in the forest, when they arrive at the wood yard \cite{Sannikov2018}, for supply-chain tracking purposes. In the context of mobile robotics, being able to uniquely identify trees would improve localization in forests \cite{Ramos2007, Smolyanskiy2017}. Robots could use trees as robust visual landmarks, in order to localize themselves. In order to perform tracking on trees, one must be able to re-identify them, potentially from bark images. In this paper, we explore this problem by developing a method to compare images of tree bark and determining if they come from the same surface or not.

\begin{figure}[!ht]
\includegraphics[width=0.48\textwidth]{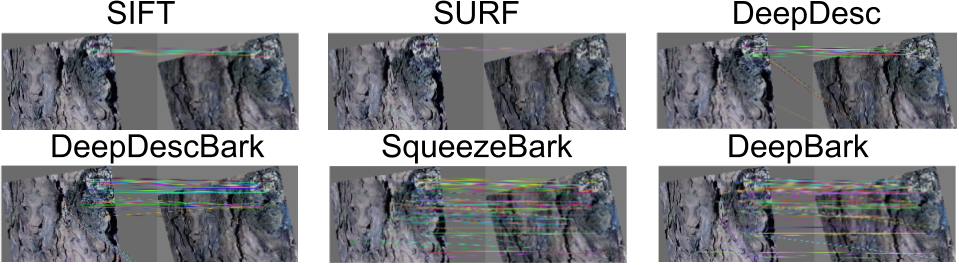}
\caption{Qualitative matching performance of descriptors, for two images of the same bark surface. Every match shown in the image passed a geometric verification. Some false positive matches still remain, due to the high level of self-similarity. Notice the strong illumination changes between the image pair, a key difficulty in tree bark re-identification.}
\label{fig:MatchingPerformance}
\vspace{0.1in}
\end{figure}

The difficulty of re-identifying bark surfaces arises in part from the self-similar nature of their texture. Moreover, the bark texture induces large changes in appearance when lit from different angles. This is due to the presence of deep troughs in the bark of many tree species. Another difficulty is the absence of a dataset tailored to this problem. There are already-existing bark datasets \cite{Fiel2010, Svab2014, Carpentier2018}, but these are geared towards tree species classification. 

To this effect, we first collected our own dataset with 200 uniquely-identified bark surface samples, for a total of 2,400 bark images. With these images, we produced a feature-matching dataset enabling the training of deep learning feature descriptors. We also have established the first state-of-the-art bark retrieval performance, showing promising results in challenging illumination conditions. In particular, it surpassed by far common local feature descriptors such as \ac{SIFT}~\cite{Lowe2004} or \ac{SURF}~\cite{Bay2006}, as well as the novel data-driven descriptor \texttt{DeepDesc}~\cite{Simo-serra2014}: see \autoref{fig:MatchingPerformance} for a qualitative assessment. 

In short, our contributions can be summarized as follows:

\begin{itemize}
  \item We introduce a novel dataset of tree bark pictures for image retrieval. These pictures also contain specific fiducial markers to infer camera plane transformation. 
  \item  Using our dataset and standard neural network architectures, we establish a new state-of-the-art performance for bark re-identification.
\end{itemize}

%%%%%%%%%%%%%%%%%%%%%%%%%%%%%%%%%%%% RELATED WORK %%%%%%%%%%%%%%%%%%%%%%%%%%%%%%%%%%%%
\section{RELATED WORK}
Our problem is related to three main areas: \emph{image retrieval}, \emph{local feature descriptors} and \emph{metric learning}, all discussed below. We also discuss the application of computer vision methods to the identification of bark images. 

\subsection{Image retrieval}
The problem of \emph{image retrieval} can be defined as follows: given a query image, we seek other images in a database that look similar to the query one. In mobile robotics, an instance of this problem is known as \ac{VPR}~\cite{Cummins2008,Cummins2009}, %,Lowry2015}, 
where image retrieval is used to perform localization. There, the objective is to determine if a location has already been visited, given its visual appearance. The robot could then localize itself by finding previously-seen and geo-referenced images. In the area of surveillance, the problem is defined as \ac{Person Re-Id}. It aims at following an individual through a number of security camera recordings~\cite{ZhengLearnedCNNEmbedding}. This implies the ability to map multiple images of an individual to the same compact description, despite variation of view-point, illumination, pose or even clothes. Our tree bark re-identification is closest to this \ac{Person Re-Id} problem, since we desire to track an individual bark surface despite changes in illumination and viewpoint.

\subsection{Local feature descriptor}
To describe and compare images while being invariant to view point and illumination changes, we based ourselves on local feature descriptors. The goal of these descriptors is to summarize the visual content of a small image patch. The ideal descriptor is \emph{a)} compact (low dimensionality) \emph{b)} fast to compute \emph{c)} distinctive and \emph{d)} robust to illumination, translation and rotations. A popular approach is to use hand-crafted descriptors. They often rely on histograms of orientation and magnitude of image gradients, as in \ac{SIFT} or \ac{SURF}.

Recently, \emph{data-driven} approaches based on machine learning have appeared~\cite{Schonberger2017}. Some learn a parametric function that maps image patches to compact descriptions that can be compared by their distance~\cite{Brown2011, Simo-serra2014}. Instead of describing an image patch alone, \cite{Han2015} takes two patches at once and directly provides a similarity probability. Some also propose a pipeline trained end-to-end (detector + descriptor)~\cite{DeTone2018SuperPointSI}.

\subsection{Metric learning}
To build a learned local feature descriptor, we turned to the field of metric learning. In this paradigm, one tries to learn a distance function between data points. More precisely, it seeks to make this distance small for similar examples, and large for dissimilar ones. This is in line with the points \emph{c)} and \emph{d)} of an ideal descriptor. This approach has been explored in \cite{Simo-serra2014, DeTone2018SuperPointSI}, where training relied on the so-called \emph{contrastive loss}. Another line of work attempts instead to make the inter-class variation larger than the intra-class variation by a chosen margin in the vector space. This formulation corresponds to the \emph{triplet loss}~\cite{Schroff2015}. \cite{Sohn2016} instead chose to compare a similar pair of examples to multiple negative ones, using a clever batch construction process.

\subsection{Vision applied to bark/wood texture}
Exploiting the information present in bark images has been explored before. For instance, hand-crafted features such as \ac{LBP} \cite{Huang2006, Svab2014}, \ac{SIFT} descriptors \cite{Fiel2010} and Gabor filters \cite{Chi2003} have been used for tree species recognition. More closely related to our work, \cite{Boudra2015ACO} compared variants of the \ac{LBP} method for image retrieval, but only at the species level. If bark is framed as a texture problem, one can find interesting work~\cite{Zhang2017HighPrecisionLU} that use ground textures such as asphalt or wood floor to enable robots to localize themselves. However, their technique is based on images with almost no variations. Moreover, each query is compared with one set of \ac{SIFT} descriptions from their whole texture map. Data-driven approaches such as deep learning also were applied on images of bark, but strictly for species classification~\cite{Carpentier2018}.

%%%%%%%%%%%%%%%%%%%%%%%%%%%%%%%%%%%% PROBLEM DEFINITION %%%%%%%%%%%%%%%%%%%%%%%%%%%%%%%%%%%%
\section{PROBLEM DEFINITION}
\begin{figure}[!ht]
\includegraphics[width=0.48\textwidth]{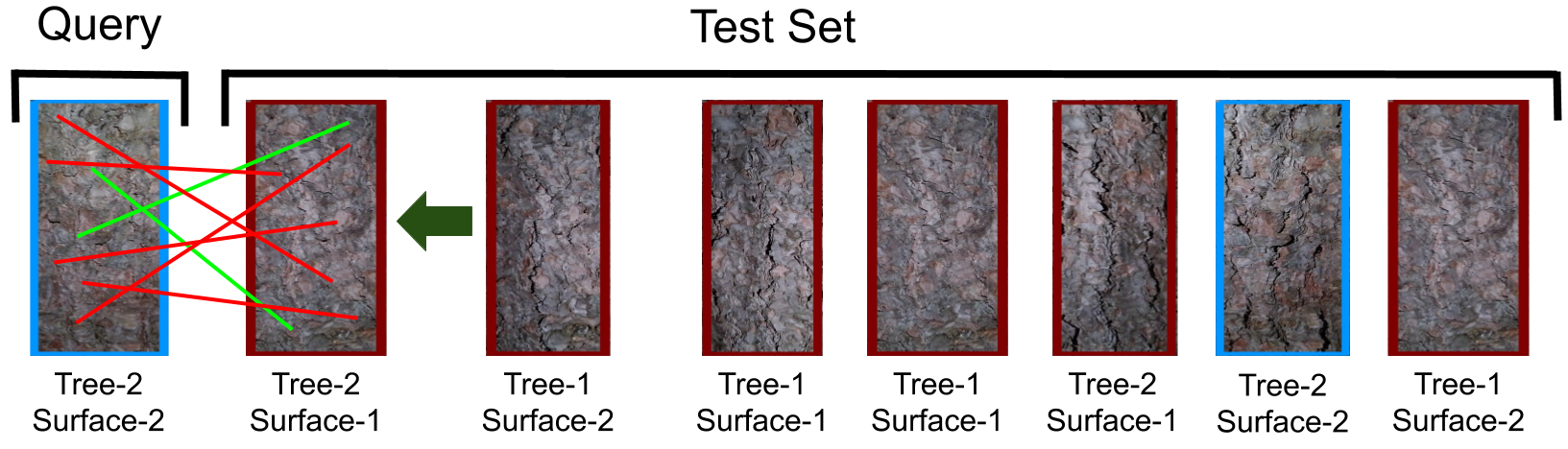}
\caption{Here is a simplified example of bark images from two different trees, on which two different surfaces have been photographed. With two images per surface, this gives us eight images. Our goal is to take one bark image and retrieve the other bark image corresponding to the exact same surface. This way, we can recognize a previously-seen tree, using bark pictures as signature.}
\label{fig:Problem_Definition}
\end{figure}

The problem we are addressing is an instance of re-identification. Given an existing database of bark images and a query image $I_q$, our goal is to find all images in the database that correspond to the \emph{same physical surface}, in order to re-identify the tree, as seen in \autoref{fig:Problem_Definition}. For instance, if the image compared to $I_q$ is from the same tree \emph{but} a different bark area, then it is not a valid match and the tree is not re-identified. We assume that $I_q$ has a meaningful match in our database, i.e., we are not solving an open-set problem; See FAB-MAP~\cite{Cummins2008} for novel locations detection.

\vspace{-5mm}

\subsection{Image global signature $s_i$}
\begin{figure}[!ht]
\centering
\includegraphics[width=0.45\textwidth]{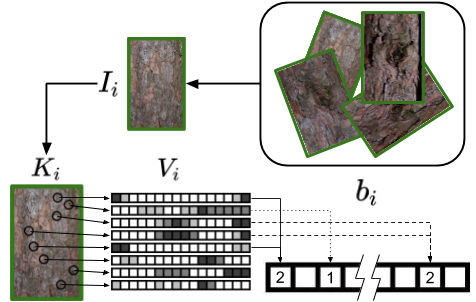}
\caption{Illustration of the global signature $s_i=(K_i, V_i, b_i)$  extraction pipeline, for a single image $I_i$. First, the keypoints $K_i$ are detected. Then, for each keypoint $k$, a descriptor $v$ is computed, creating the list $V_i$. Finally, a \acf{BoW} representation $b_i$ of $V_i$ is computed from the quantization of all descriptions $v$ via a visual vocabulary.}
\label{fig:SignDataset}
\vspace{2mm}
\end{figure}

We perform the bark image search via global image signatures, defined as $s_i=(K_i, V_i, b_i)$. These signatures are extracted for each image (database and query $I_q$), as depicted in \autoref{fig:SignDataset}. For this, we mostly follow the method used in~\cite{Sivic2003}, summarized below. First, a keypoint detector extracts a collection $K_i$ of keypoints from an image. For each of these keypoints $k \in K_i$, we extract a description $v$ of dimension 128, yielding a list of descriptions $V_i$. These descriptions can be from standard descriptors, such as \ac{SIFT} or \ac{SURF}, or our novel descriptors, described further below. The remaining component of an image signature $s_i$ is a \ac{BoW} representation $b_i \in \mathbb{R}^{1000}$, calculated from the list of descriptions $V_i$. We also apply the standard \emph{TF-IDF} technique. In~\cite{Sivic2003}, the comparison between two \ac{BoW} is done using the cosine distance. Instead, we have $l_2$-normalized once every \ac{BoW} as a pre-processing step and use the $l_2^2$ distance to compare them. This way, our distance ranking is equivalent to the pure cosine distance, but without using a dot product.

\subsection{Signature matching}
The search is performed mainly by computing a score \Score{} between a query image signature $s_q$ and each image signature $s_i$ in the database, and retrieving the best match based on \Score{}. For the \ac{BoW} technique, we simply use the distance between two \ac{BoW}s $||b_q - b_i||^2_2$ as our score \Score{}. Another way to calculate a score between $s_q$ and $s_i$ begins by taking the $l_2^2$ distance between every description of $V_q$ and $V_i$ to obtain a collection $M$ of putative matching pairs of features, ${m \in M =(v \in V_q,v \in V_i)}$ with ${|M|=|V_q|}$. Then, we explored the use of two potential false match filters. The first one is the \emph{\ac{LR} test} introduced in~\cite{Lowe2004}. The second one is a \ac{GV}, which is a simple neighbors check. It begins by taking a match $m=(v_x, v_y)$, then retrieving the keypoints $(k_x, k_y)$ associated with each description $v$ of the match. Following this, we find the \NbNearestNeighbor{} nearest neighbors of each of the keypoints in their respective images. Finally, the match is accepted if \emph{at least} \NearestNeighborMatchingRatio{}\% of the \NbNearestNeighbor{} neighbors of $k_x$ have a match $m \in M$ with the \NbNearestNeighbor{} neighbors of $k_y$. The number of matches after filtering is then considered as the matching score \Score{}.

\section{Our approach: Data-driven descriptors}
Considering that tree bark highly-textured surfaces are problematic to hand-crafted descriptors, and the non-availability of datasets tailored to our task (see \autoref{tab:Existing_Bark_Dataset}), we present here the main contribution of our paper, which is data-driven descriptors for bark image re-identification. First, we describe our novel bark image dataset. We then discuss how to process it in order to generate keypoint-aligned image patches required to train our descriptors. These descriptors are then described in detail, followed by training details.

\begin{table}[!ht]
\centering
\begin{tabular}{| c || c | c | c | c |}

 \hline
 Reference                      & \makecell{Number of \\ images}  &  Public   & \makecell{Instance \\ Retrieval} & \makecell{Pixel \\ Align}\\
 \hline
 \hline
 \cite{Chi2003}                                  & 200               &         &     &   \\
 \hline             
 \cite{Huang2006}                               & 300               &         &     &  \\
 \hline             
 TRUNK12, \cite{Svab2014}                       & 393               &  \checkmark      &     &  \\
 \hline     
 \cite{Bressane2015}                             & 540               &         &     &  \\
 \hline     
 \cite{Blaanco2016}                             & 920               &         &     &  \\
 \hline             
 AFF, \cite{Fiel2010}                           & 1183              &         &     &  \\
 \hline 
 BarkNet, \cite{Carpentier2018}                 & 23616             &  \checkmark      &     &  \\
 \hline 
 Ours                                        & 2400+750          &  \checkmark      & \checkmark   & \checkmark \\
 \hline
\end{tabular}
\caption{Comparative of the existing bark datasets based on their size, availability, and applications. All datasets contain tree bark images designed for species classification, except ours. Adapted from \cite{Carpentier2018}.}
\label{tab:Existing_Bark_Dataset}
\end{table}
\vspace{-4mm}

\subsection{Bark Image Datasets}\label{sec:database}
\begin{figure}[!ht]
\centering
\includegraphics[width=0.35\textwidth]{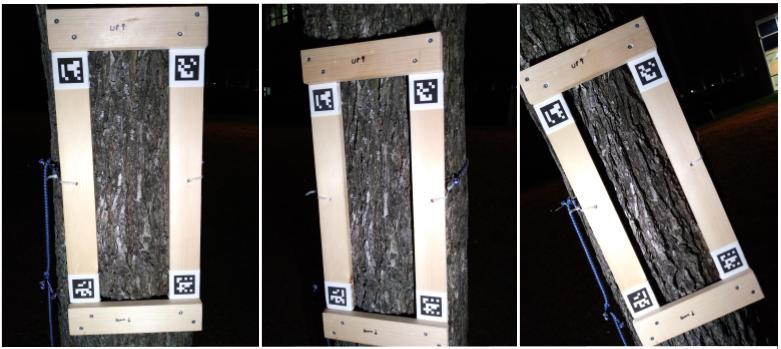}
\caption{Images from our database of the same surface of \acf{EL} bark, but for different illuminations and camera angles. In each image, there are four fiduciary markers on a custom-made wooden frame, used for pixel-wise registration.}
\label{fig:BarkExample}
\end{figure}

In order to develop our data-driven descriptors, we collected a dataset of tree bark images. To ensure drastic illumination changes, we took the pictures at night, and varied the position of a 550 lumen LED $Energizer^{TM}$ lamp. We also varied the position of the camera, an LG Q6 cellphone with a resolution of $4160 \times 3120$ pixels. Since our training approach (\autoref{sec:DataDrivenDataset}) requires keypoint-aligned image patches, we used fiducial markers on a wooden frame attached to trees to automate and increase the precision of the image registration, as shown in \autoref{fig:BarkExample}. 

We collected bark images for two different tree species, namely \ac{RP} (an evergreen) and \ac{EL} (a deciduous tree). For each species, 50 trees were selected, on which we further chose two different and distinct surfaces and took 12 photos for each of these surfaces. Each bark was surrounded by a custom-made wooden frame of 50.5~$cm$ by 15~$cm$. We limited ourselves to only two species, to avoid positively biasing image retrieval results. Indeed, neural networks have the capacity to easily distinguish between tree species~\cite{Carpentier2018}. In total, we took 12 images per distinct surface with the aforementioned variations. To make our evaluation on \ac{EL} bark more challenging, we also collected unseen bark images from elm trees without any markers. To keep these new images close to our original appearance distribution, we took them at night with three different illumination angles, but with limited changes in point of view. We collected a total of 30 images per tree for 25 trees with some physical overlap, spread nearly uniformly around the trunk. This gave us a total of \nbNegTree{} manually-cropped non-relevant images for any \ac{EL} query taken at a scale similar to all of our other images.

\vspace{-5mm}

%%%%%%%%%%%%%%%%%% LEARNED DESCRIPTOR %%%%%%%%%%%%%%%%%%%%%
\subsection{Descriptor Training Dataset}\label{sec:DataDrivenDataset}
\begin{figure}[!ht]
\centering
\includegraphics[width=0.42\textwidth]{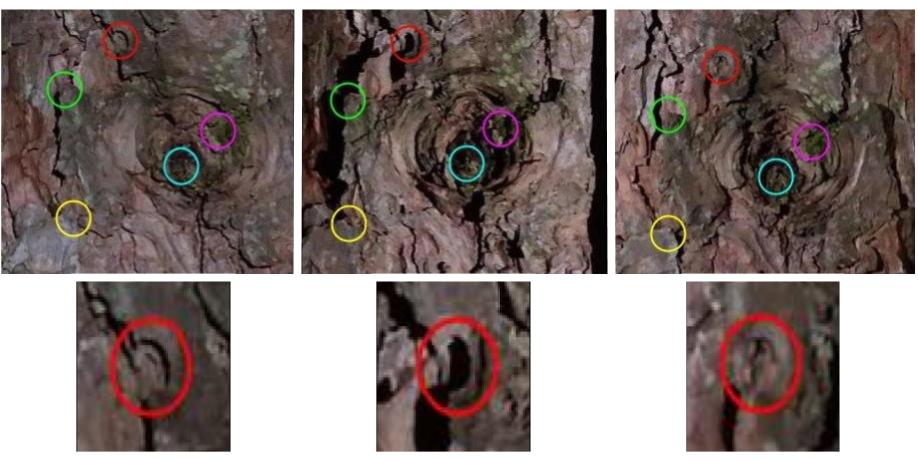}
\caption{Top row: pictures of the same bark surface with strong changes in illumination. Each circle color is a distinct keypoint. Bottom row: close up of the red keypoints from their respective images. This highlights the importance for a descriptor to be as immune as possible to such illumination changes.}
\label{fig:BarkFeature}
\end{figure}

\begin{figure}[!ht]
\centering
\includegraphics[width=0.42\textwidth]{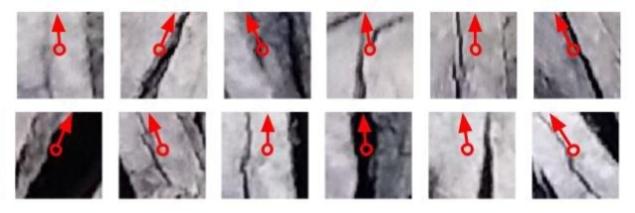}
\caption{Actual example of $64 \times 64$ patches of a keypoint. Red arrows indicate the orientation of the original bark images.}
\label{fig:FeatureWithOrientation}
\end{figure}

Our descriptors require a dataset of $64 \times 64$ patches for training with metric learning. Moreover, these patches not only need to be properly indexed per bark surface, they must also be centered around the same physical location, corresponding to the keypoint. After automatically cropping the excess information from images (background, frame, shadow, etc), we performed registration between every image of a bark surface with a reference frame \RefFrame{} via a homography \HomoImgToRef{}. We used the fiducial markers affixed to our wooden frame surrounding the bark surface (See \autoref{fig:BarkExample}) to estimate these transformations. Then, for each bark image, we detected the maximum number of keypoints and projected them to the reference frame \RefFrame{} via the homography \HomoImgToRef{}. We filtered all of the keypoints in \RefFrame{} to require a minimum distance of 32 pixels between them to minimize overlap. This resulted in around 800-1000 distinct keypoints in \RefFrame{}. For each of these keypoints, we then found the 12 image patches (one per image, see~\autoref{sec:database}) using a homography \HomoRefToImg{} that gives the transformation from the reference frame \RefFrame{} to a specific bark image. This resulted in a collection of $64 \times 64$ image patches corresponding to the exact same physical location on the bark, but with changes in illumination and point of view (rotation, scaling and perspective). \autoref{fig:BarkFeature} shows three images of a unique bark surface, with the manual correspondence between keypoints. \autoref{fig:FeatureWithOrientation} shows 12 examples of a keypoint extracted according to our algorithm used to create the training dataset. 

\vspace{-5mm}

\vspace{-5mm}

\subsection{DeepBark and SqueezeBark Descriptors}\label{sec:OurDescriptors}
To perform description extraction, we implemented two different architectures with Pytorch 0.4.1. The first one, \texttt{DeepBark}, is based on a pre-trained version of \mbox{ResNet-18} on ImageNet. We removed the average pool and the fully connected layers and replaced them with one fully connected layer (no activation function). The second one, \texttt{SqueezeBark}, is a smaller network based on the pre-trained version of SqueezeNet 1.1~\cite{Iandola2016} on ImageNet. We again removed the average pool and the fully connected layers. We replaced them with a max pooling layer (to reduce the feature map) and a fully connected layer (no activation function). In both cases, the network computes a 128-dimensional vector, fed to an $l_2$ normalization layer. Removing our last fully connected layer and calculating the number of parameters for the remaining convolutionnal layers, \texttt{DeepBark} is then composed of a total of 10,994,880 parameters and \texttt{SqueezeBark} includes 719,552 parameters. Our intention here is to be able to compare a number of network representation powers on the descriptor quality.

\subsection{Training details}
To train our networks (\texttt{DeepBark} or \texttt{SqueezeBark}), we chose the N-pair-mc loss~\cite{Sohn2016}. The only difference in our implementation is that, instead of using $l2$ regularization to avoid degeneracy, we $l2$-normalized the descriptor vectors \DescVect{} to keep them in a hypersphere~\cite{Schroff2015}. 

Our dataset is composed of $64 \times 64$ patches around 70,800 distinct keypoints for the training set and 17,500 for the validation set for most of our experiments. Using 12 patches by keypoint for training and 2 for validation, this totals 884,600 $64 \times 64$ bark images patches. At each iteration, we only used a pair of examples for every keypoint in the training set. However, to ensure an equal probability for every patch to be seen together with every other patch, we randomly selected each patch tuple. We added online data augmentation in the form of color, luminosity and blurriness jitter. Each input image was normalized between $(-1, 1)$ by subtracting 127.5 and then divided by 128. We optimized using Adam starting with a learning rate of $1e^{-4}$ and reducing it by a factor of 0.5 each time the validation plateaued for 20 iterations.

We built the validation set by finding all keypoints in the bark images set aside for validation, and randomly selected 2 patches from the 12 available for each distinct keypoint. This gave us a fixed validation set, where every patch had a corresponding one. During training we validated our model by selecting 50 keypoints with their 2 examples at the time and performed a retrieval test to calculate the \ac{P@1}. The final validation score was simply the average of every \ac{P@1} calculated for every batch of 50 keypoints. After training, we selected the model with the highest validation score. The training was stopped either with early stopping when the validation stagnated for 40 iterations, or when a maximum number of iterations was reached.

%%%%%%%%%%%%%%%%%%%% RESULTS %%%%%%%%%%%%%%%%%%%%%%%%
\section{RESULTS}
Together with \texttt{DeepBark} and \texttt{SqueezeBark}, we evaluated hand-crafted descriptors, namely \ac{SIFT} and \ac{SURF}. We also included \texttt{DeepDesc}, a learned descriptor originally trained on the multi-view stereo dataset~\cite{Brown2011} and our re-implementation of \texttt{DeepDesc} renamed \texttt{DeepDescBark}, which we train on bark data following our training procedure. All descriptors used the \ac{SIFT} keypoint detector, except for \ac{SURF} that used its own detector. For all experiments, we used a ratio of 0.8 for the \ac{LR} test, and set $\alpha=15$ and $\rho=0.33$ for the \ac{GV} filter. Also, each visual vocabulary $voc$ was computed from the training images of each respective experiment, while being clustered using the $k$-mean algorithm.

Image retrieval can be evaluated in multiple ways. In our case, we favored metrics based on an ordered set, as they align best with our problem. Here are \ac{P@K} and \ac{R@K}:
\vspace{2mm}
\noindent\begin{minipage}{.5\linewidth}
\begin{equation}\label{eq:P@K}
  P@K=\frac{p(K)}{K},
\end{equation}
\end{minipage}%
\begin{minipage}{.5\linewidth}
\begin{equation}\label{eq:R@K}
  R@K=\frac{p(K)}{|I|}.
\end{equation}
\end{minipage}\vspace{2mm}
In \autoref{eq:P@K} and \ref{eq:R@K}, $K$ is a rank and the function $p$ returns the number of relevant images ranked between the first rank and the $K$ rank ($K$ included). For \autoref{eq:R@K}, $I$ is the set of relevant images. We also present \ac{PR} graph defined as:
\begin{equation}\label{eq:PR}
  PR=\{i \in I \mid(R@i_k,P@i_k)\}.
\end{equation}
In \autoref{eq:PR}, $i \in I$ represents one image and $i_k$ is the rank where the image $i$ can be found. Taking the mean of every $P@i_k$ gives the \ac{AP}. Keep in mind that these metrics are calculated for every query, averaged together. Thus, instead of \ac{AP} we write \ac{mAP}.

\subsection{Hyperparameters search}
Our approach comprises three hyperparameters: \NbKp{}, \Downsize{} and  \BlurSigma{}. The first one, \NbKp{}, is the maximum number of keypoints in an image. From experiments, increasing \NbKp{} beyond 500 did not significantly improve the performance of any descriptor. The second hyperparameter is the downsizing factor \Downsize{} of the original image. Downsizing allowed the receptive field of any method to be increased, without changing its process. Our experiments showed that using \Downsize{} $=2$ generally helped every descriptor. Our third hyperparameter \BlurSigma{} is the size of the gaussian kernel used in the blurring performed before passing the image through the keypoint detector. Note that the blur was used for the keypoint detection, but after that we used either the unblurred image to compute the description of learned descriptors (\texttt{DeepBark}, \texttt{SqueezeBark}, \texttt{DeepDesc} and \texttt{DeepDescBark}) or the blurred image for \ac{SIFT} and \ac{SURF}. The latter was necessary, as they used the keypoint information found on the blurred image. We found that the best blur filter value \BlurSigma{} varied greatly between descriptors. The chosen values for the \autoref{sec:ComparativeExp} experiment are shown in \autoref{tab:Chosen_Hyper_param}. These values were found by averaging the results over 36 randomly-selected queries run on the validation set for each hyperparameter combination.

\begin{table}[!ht]
\centering
\begin{tabular}{| c | c | c | c | c |}

 \hline
 Descriptors            & \Downsize{} & \BlurSigma{} & \ac{mAP} & Avg. Keypoint Num. \\
 \hline
 \hline
 \ac{SIFT}              & 1.5   & 3 & 0.406 & 469.4 $\pm$69.9       \\
 \hline
 \ac{SURF}              & 2.0   & 3 & 0.489 & 499.6 $\pm$4.8 \hfill \\
 \hline
 \texttt{DeepDesc}      & 2.0   & 1 & 0.091 & 497.0 $\pm$17.4       \\
 \hline
 \texttt{DeepDescBark}  & 2.0   & 0 & 0.076 & 492.8 $\pm$18.4       \\
 \hline
 \texttt{SqueezeBark}   & 2.0   & 0 & 0.183 & 492.8 $\pm$18.4       \\
 \hline
 \texttt{DeepBark}      & 2.0   & 0 & 0.966 & 492.8 $\pm$18.4       \\
 \hline

\end{tabular}
\caption{Hyperparameters chosen after careful examination of the grid search, with the mean number of keypoints found at test time. The number of keypoints was capped to 500. \ac{mAP} results shown have been obtained with the \ac{GV} method. Some descriptors perform better with other scoring methods not shown here.}
\label{tab:Chosen_Hyper_param}
\end{table}
\vspace{-2mm}

\subsection{Impact of training data size}
Data-driven approaches based on Deep Learning tend to be data hungry. To check the impact of the training data size, we created 5 training scenarios by tree species, which used 10\%, 20\%, 30\%, 40\% and 50\% of the dataset. All trained descriptors were validated and tested on the same set (10\% and 40\% respectively) of each species dataset. We stopped training when the validation \ac{P@1} stagnated for 40 consecutive iterations. 

\autoref{tab:trainingSaturation} shows the performance of the descriptor \texttt{DeepBark}, for each training set size. For each species, the \ac{P@1}, the \ac{R-P} and the \ac{mAP} are reported for the three scoring techniques: \ac{GV}, \ac{LR} and \ac{BoW}. It is good to note that the \ac{BoW} is also affected by the size of the training set, since the $voc$ of the \ac{BoW} is computed from that same training set. From these metrics, we concluded that performance gains were minimal beyond 40\%. This confirmed that our training database is sufficiently large to obtain good performance. For references, when using 50\% of \ac{RP} as training data, we have access to approximately 42,700 distinct keypoints giving 512,000 bark image patches of $64 \times 64$ pixels.

\begin{table}[!ht]
\centering
\begin{tabular}{| @{} l | c || c | c | c | c | c |}

 \hline
 \multicolumn{7}{|c|}{Red Pine} \\
 \hline
 
                       & Metric    & 10\%  & 20\%  & 30\%  & 40\%  & 50\% \\ 
 
 \hline
 
 \multirow{3}{15pt}{\ac{BoW}}& \ac{P@1}  & 0.971 & 0.985 & 0.985 & \textbf{0.996} & 0.994 \\
 \cline{2-7}
                            & \ac{mAP}  & 0.633 & 0.713 & 0.769 & 0.785 & \textbf{0.812} \\
 \hline
 
 \multirow{3}{15pt}{\ac{GV}} & \ac{P@1}  & 0.988 & 0.990 & \textbf{0.998} & 0.996 & \textbf{0.998} \\
 \cline{2-7}
                            & \ac{mAP}  & 0.777 & 0.848 & 0.892 & 0.905 & \textbf{0.922} \\
 \hline
 
 \multirow{3}{15pt}{\ac{LR}} & \ac{P@1}  & \textbf{1.000} & \textbf{1.000} & \textbf{1.000} & \textbf{1.000} & \textbf{1.000} \\
 \cline{2-7}
                            & \ac{mAP}  & 0.882 & 0.932 & 0.956 & 0.962 & \textbf{0.967} \\
 \hline
 \hline
 \multicolumn{7}{|c|}{Elm} \\
 \hline
 
                      & Metric    & 10\%  & 20\%  & 30\%  & 40\%  & 50\% \\ 
 
 \hline
 
 \multirow{3}{15pt}{\ac{BoW}}& \ac{P@1}  & 0.940 & 0.956 & 0.971 & 0.979 & \textbf{0.983} \\
 \cline{2-7}
                            & \ac{mAP}  & 0.607 & 0.691 & 0.721 & 0.759 & \textbf{0.764} \\
 \hline
 
 \multirow{3}{15pt}{\ac{GV}} & \ac{P@1}  & 0.944 & 0.965 & 0.977 & 0.981 & \textbf{0.983} \\
 \cline{2-7}
                            & \ac{mAP}  & 0.707 & 0.752 & 0.791 & \textbf{0.816} & 0.806 \\
 \hline
 
 \multirow{3}{15pt}{\ac{LR}} & \ac{P@1}  & 0.985 & 0.996 & 0.998 & 0.998 & \textbf{1.000} \\
 \cline{2-7}
                            & \ac{mAP}  & 0.665 & 0.740 & 0.779 & \textbf{0.800} & 0.798 \\
 \hline
 
\end{tabular}
\caption{Performance of the \texttt{DeepBark} descriptor, when training with 10\%, 20\%, 30\%, 40\% and 50\% of the data from a single tree species. The remaining data has been used for validation (10\%) and testing (40\%). Hyperparameters were fixed through testing. Best results are in bold for each row.}
\label{tab:trainingSaturation}
\vspace{-6mm}
\end{table}

\subsection{Descriptors comparison}\label{sec:ComparativeExp}
\begin{figure*}[!ht]
\includegraphics[width=0.98\textwidth]{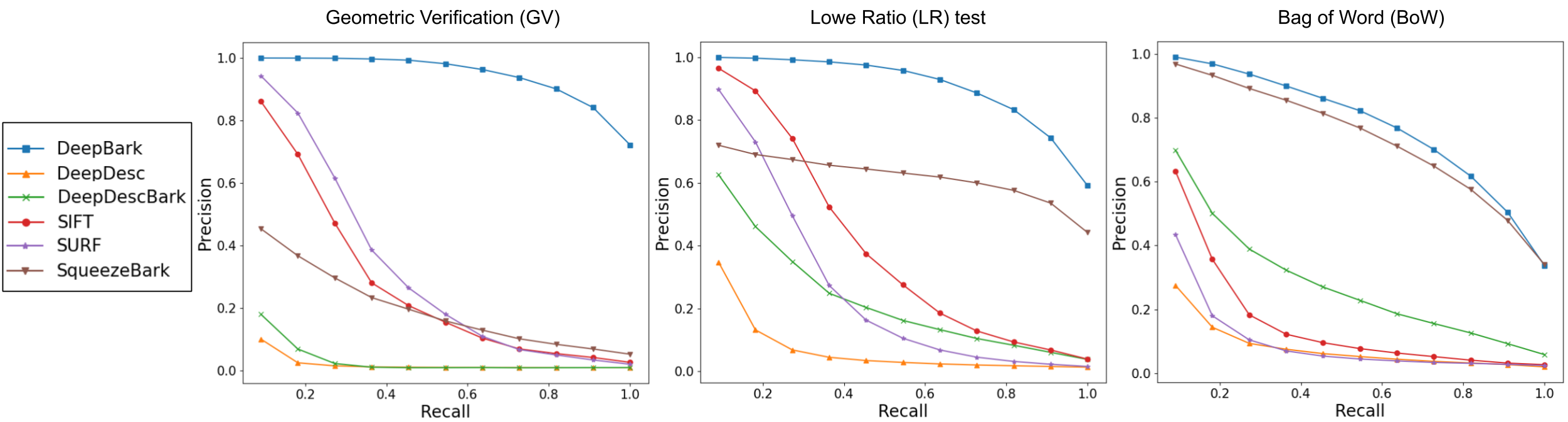}
\caption{\ac{PR} Curve for all descriptors tested on 50\% of \ac{RP} and \ac{EL}. Learned descriptors were trained on the remaining 50\% of bark. Each of the 1200 images of the test set is use as a query. No extra negative examples were added.}
\label{fig:Test_5050_RP-EL_AP_Curve}
\vspace{-2mm}
\end{figure*}

We selected 50\% of red pine bark surfaces and 50\% of elm bark surfaces to create a test set, while using the remaining data for the training and validation sets. This corresponded to 80 unique bark surfaces for the training, 20 for the validation and 100 for testing, while keeping the ratio between tree species to 50/50 in each set. The data-driven descriptors  \texttt{DeepBark}, \texttt{SqueezeBark} and \texttt{DeepDescBark} were trained for 200 iterations, and we kept their model with the best validation. With 12 images for each bark surface, the test set had a total of 1200 images, with 600 per tree species. Each of these images was used as a query during the retrieval test. The results were averaged over all queries. We report results in \autoref{fig:Test_5050_RP-EL_AP_Curve} as \ac{PR} curves. This way, all 11 true positives are taken into account in our experimentation, properly estimating how well our approach resists to strong illumination/viewpoint changes. 

From \autoref{fig:Test_5050_RP-EL_AP_Curve}, we can see that hand-crafted descriptors often successfully retrieve one image, but struggle beyond this. We can also see that \texttt{DeepBark} clearly dominates all descriptors. We can also notice that the precision is over 98\% up to a recall of 6 images, when it is combined with \ac{GV}. Interestingly, the results for \texttt{SqueezeBark} are mitigated. This might indicate that finding a good descriptor for bark images under strong illumination changes is a difficult problem, requiring a neural architecture with sufficient capacity. This is further supported by \texttt{DeepDescBark} exhibiting worse performance than \texttt{SqueezeBark} and \texttt{DeepBark}, which are larger networks. The very-low capacity of \texttt{DeepDesc} might also explain why it performed worse that \ac{SIFT} or \ac{SURF} at times. Finally, one can see that appropriate data improves performance, as demonstrated by the performance of \texttt{DeepDescBark} over \texttt{DeepDesc}.

\subsection{Generalization across species}
\begin{figure}[!ht]
\centering
\includegraphics[width=0.42\textwidth]{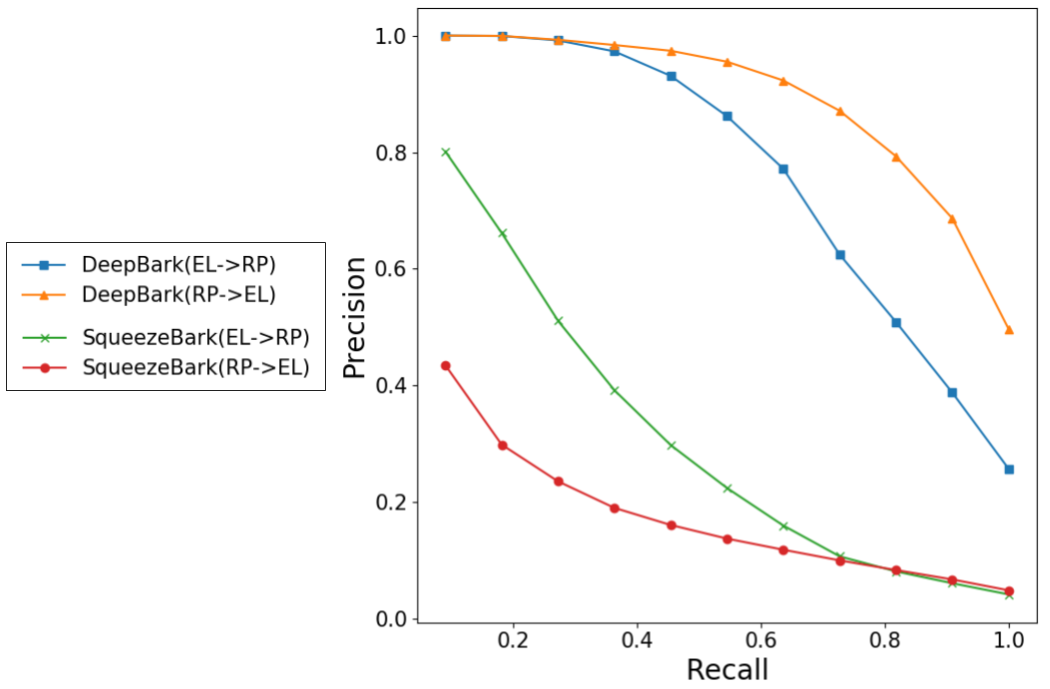}
\caption{\ac{PR} curve for the generalization test using the \ac{GV} method. The arrow -> indicates the generalization direction (trained on -> tested on).}
\label{fig:Test_Generalization_AP_Curve}
\end{figure}

In the experiments of \autoref{sec:ComparativeExp}, we reported results on networks trained on both species, instead of training and testing each architecture on a single species. Our intention was to double the amount of training data and benefit from the potential synergy between species, which is often seen in deep networks (multi-task learning). Here, we precisely look at the generalization of our networks across species. We thus devised two experiments to evaluate the generalization from one species to the other and vice versa. The first one is composed of a training set with 80\% of the \ac{RP} data, using the remaining 20\% as the validation set and all of the \ac{EL} data as the test set (labelled \ac{RP}->\ac{EL}). We also performed the converse (\ac{EL}->\ac{RP}). We only report in \autoref{fig:Test_Generalization_AP_Curve} the \ac{PR} curve for the \ac{GV}, as the trend is similar for other scoring methods. \autoref{fig:Test_Generalization_AP_Curve} first shows that \texttt{DeepBark} is capable of generalizing across species, but that \texttt{SqueezeBark} do so to a lesser extent. Also, there is no clear trend for the generalization direction, since \texttt{SqueezeBark} generalized better from \ac{EL} to \ac{RP} but \texttt{DeepBark} generalized better in the opposite direction (from \ac{RP} to \ac{EL}). 

\subsection{Extra negative examples}
To extrapolate how our system would perform on a larger database, we added 6,700 true negative elm examples with a crop size similar to query images. Half of them were original images, and the other images were generated via data augmentation, by doing either a rotation, scale or affine transformation. Note that the original 3,350 images contain some physical overlap, as they come from 25 trees. 

\begin{figure*}[!ht]
\includegraphics[width=0.98\textwidth]{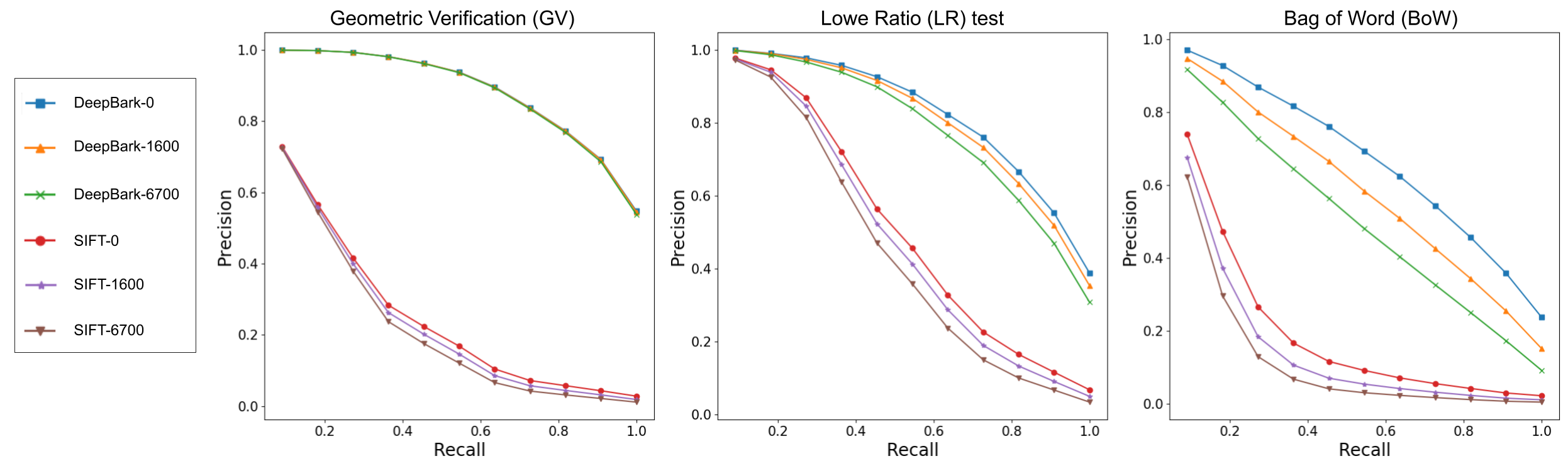}
\caption{\ac{PR} Curve for the negative examples test on \ac{SIFT} and \texttt{DeepBark}. The number in the legend indicates how many negative examples were added.}
\label{fig:NegEx_AP_Curve}
\vspace{-5mm}
\end{figure*}

We reused the \texttt{DeepBark} network and the $voc$ previously trained in \autoref{sec:ComparativeExp}. For the test, we removed the red pine images and kept the elm images that we separated into two crops (top and bottom halves) giving us a total of 1,200 images. We thus obtain a database of 7,900 bark images. Again, every query had 11 relevant images. This experiment is the only one where we split bark images into two crops, solely done to increase the database size. This has a negative impact on the performance, as the visible bark (and thus the number of visible features) is reduced by half. This can be seen by comparing \autoref{fig:Test_5050_RP-EL_AP_Curve} and \autoref{fig:NegEx_AP_Curve}.

\begin{table}[!ht]
\centering
\begin{tabular}{| @{} l | c || c | c | c |}

 \hline
                                & Metric    & 0                         & 1600                      & 6700 \\ 
 
 \hline
 \hline
 
 \multirow{3}{15pt}{\ac{BoW}}   & \ac{P@1}  & \textbf{0.952 $\pm$0.214} & 0.924 $\pm$0.265          & 0.885 $\pm$0.319 \\
 \cline{2-5}
                                & \ac{mAP}  & \textbf{0.659 $\pm$0.258} & 0.572 $\pm$0.276          & 0.491 $\pm$0.281 \\
 \hline
 
 \multirow{3}{15pt}{\ac{GV}}    & \ac{P@1}  & \textbf{0.998 $\pm$0.041} & \textbf{0.998 $\pm$0.041} & \textbf{0.998 $\pm$0.041} \\
 \cline{2-5}
                                & \ac{mAP}  & \textbf{0.874 $\pm$0.165} & \textbf{0.874 $\pm$0.166} & 0.872 $\pm$0.168 \\
 \hline
 
\multirow{3}{15pt}{\ac{LR}}     & \ac{P@1}  & \textbf{0.999 $\pm$0.029} & 0.998 $\pm$0.050          & 0.998 $\pm$0.050 \\
 \cline{2-5}
                                & \ac{mAP}  & \textbf{0.812 $\pm$0.198} & 0.794 $\pm$0.209          & 0.768 $\pm$0.223 \\
 \hline
 
\end{tabular}
\caption{Results of the negative examples test for \texttt{DeepBark}. The number in the header indicates how many negative examples were added. The best results for each metric are in bold.}
\label{tab:Negative_Examples_Test}
\end{table}
\vspace{-2mm}

Among the three scoring methods evaluated, the most affected by the amount of negative examples was the \ac{BoW}, as seen in \autoref{fig:NegEx_AP_Curve} and \autoref{tab:Negative_Examples_Test}. The \ac{LR} filter displays a smaller degree of degradation, as a function of the amount of extra negative examples. However, it still retains almost the same \ac{P@1}. Finally, when looking at the \ac{GV}, it is clear that the impact of extra negative examples is negligible. This again demonstrates the importance of performing \ac{GV} filtering. We can thus conclude that our approach with \ac{GV} would work on a much larger, realistic dataset.

\subsection{Computing time considerations}

\begin{table}[!ht]
\centering
\begin{tabular}{| c | c | c | c | c |}

 \hline
 Descriptors    & R@25  & R@50  & R@100 & R@200 \\
 \hline
 \hline
 \ac{SIFT}-0    & 0.248 & 0.316 & 0.403 & 0.520 \\
 \hline
 \ac{SIFT}-6700 & 0.150 & 0.176 & 0.215 & 0.268 \\
 \hline
 DeepBark-0     & 0.728 & 0.795 & 0.857 & 0.908 \\
 \hline
 DeepBark-6700  & 0.561 & 0.625 & 0.681 & 0.739 \\
 \hline

\end{tabular}
\caption{\ac{R@K} for different values of K using the \ac{BoW}. Results taken from the experiment with negative examples. The number beside the method names indicates how many negative examples were added.}
\label{tab:Bow_Recall}
\end{table}
\vspace{-5mm}

Even if the \ac{LR} test and the \ac{GV} filter perform better, it is unrealistic to use them to search a whole database. Instead, the \ac{BoW} can be used as pre-filtering to propose putative candidates to the other methods. To this end, we provide \autoref{tab:Bow_Recall}, which shows the \ac{R@K} for various $K$ for the \ac{BoW} approach. These results suggest that keeping the 200 best matching scores calculated using the \ac{BoW} on \texttt{DeepBark} would retain 73.9\% of the 11 relevant images among 7,900 possible matches. As shown by \cite{Cummins2009}, the \ac{BoW} is fast to compare and can handle large datasets. To get a sense of the time that could be saved by the pre-filtering, we calculated the averaged time of 500 signature comparisons using our actual algorithms on a single thread of an Intel Core I-7. The \ac{BoW}, \ac{GV}, \ac{LR} methods took respectively 0.002, 131.5 and 179.4 $ms$ on averaged. It is important to note that the \ac{BoW} technique could be sped up further using an inverted index and by taking advantage of its sparsity (on average 71.8\% of it has a null entry in our experiments). From this, we can see that applying the \ac{GV} on the $K=200$ top from the original 7,900 images can be accomplished in 35.88~$s$, while the \ac{BoW} only took 0.016~$s$ for the 7,900 images.

%%%%%%%%%%%%%%%%%%%%%%%%%%%%%%%%%%%% CONCLUSION %%%%%%%%%%%%%%%%%%%%%%%%%%%%%%%%%%%%
\section{CONCLUSION}
In this paper, we explored bark image re-identification in the challenging context of strong illumination and viewpoint variations. To this effect, we introduced a novel bark image dataset, from which we can extract over 2 million keypoint-registered image patches. Using the latter, we developed two local feature descriptors based on Deep Learning and metric learning, namely \texttt{DeepBark} and \texttt{SqueezeBark}. We showed that both our descriptors performed better than \ac{SIFT}, \ac{SURF} and \texttt{DeepDesc} on any of the three scoring methods presented. Our results indicate that using our descriptor \texttt{DeepBark}, retrieval is viable even for large datasets with thousands of negative examples. Moreover, the approach can be sped up by using Bag-of-Words.

Our results are encouraging, but performance in a real-life scenario might differ. More data should be collected, in particular we should expand collection and testing on more tree species, as we have only tested generalization across one species to another. Also, it would be interesting to quantify the effect of the \ac{BoW} size, the generalization capacity over more tree species or the effect of using other keypoint detectors. The training procedure could be further improved, in allowing for longer training, testing other networks, adding pre-training or using hard mining approaches.

%%%%%%%%%%%%%%%%%%%%%%%%%%%%%%%%%%%% ACKNOWLEDGEMENT %%%%%%%%%%%%%%%%%%%%%%%%%%%%%%%%%%%%
\section{ACKNOWLEDGEMENT}
We would like to thank Marc-Andr\'e Fallu for the access to his red pine plantation. The authors would also like to thank the "Fonds de recherche du Québec – Nature et technologies (FRQNT)" for their financial support. We are also grateful to Fan Zhou for his help in the data collection. Finally, we thank NVIDIA for their Hardware Grant Program.

\addtolength{\textheight}{-2cm}

% trigger a \newpage just before the given reference
% number - used to balance the columns on the last page
% adjust value as needed - may need to be readjusted if
% the document is modified later
%\IEEEtriggeratref{8}
% The "triggered" command can be changed if desired:
%\IEEEtriggercmd{\enlargethispage{-5in}}

% references section

% can use a bibliography generated by BibTeX as a .bbl file
% BibTeX documentation can be easily obtained at:
% http://www.ctan.org/tex-archive/biblio/bibtex/contrib/doc/
% The IEEEtran BibTeX style support page is at:
% http://www.michaelshell.org/tex/ieeetran/bibtex/
%\bibliographystyle{IEEEtran}
% argument is your BibTeX string definitions and bibliography database(s)
%\bibliography{IEEEabrv,../bib/paper}
%
% <OR> manually copy in the resultant .bbl file
% set second argument of \begin to the number of references
% (used to reserve space for the reference number labels box)
%%%%%%%%%%%%%%%%%%%%%%%%%%%%%%%%%%%%%%%%%%%%%%%%%%%%%%%%%%%%%%%%%%%%%%%%%%%%%%%%
\bibliography{main}

% that's all folks
\end{document}